\documentclass[a4paper,fleqn]{cas-dc}
\usepackage{amssymb}
\usepackage{hyperref}
\usepackage{graphicx}
\usepackage{lipsum}
\usepackage{lscape}
\usepackage{amsmath}
\usepackage{relsize}%
\usepackage{multirow}
\usepackage[figuresright]{rotating}
\usepackage{lineno}
\usepackage{algorithm} 
\usepackage{algpseudocode} 
\usepackage{comment}
\usepackage{lineno}
\usepackage{nomencl}
\makenomenclature
\usepackage{calc}
\usepackage[T1]{fontenc}
\usepackage{relsize}
\usepackage{subfigure}

\usepackage{natbib}
\setcitestyle{authoryear,open={(},close={)},citesep={;}} 

\hypersetup{
  colorlinks,
  citecolor=Violet,
  linkcolor=Red,
  urlcolor=Blue}

\newcolumntype{P}[1]{>{\centering\arraybackslash}p{#1}}

\def\tsc#1{\csdef{#1}{\textsc{\lowercase{#1}}\xspace}}
\tsc{WGM}
\tsc{QE}
\tsc{EP}
\tsc{PMS}
\tsc{BEC}
\tsc{DE}

\begin{document}
\let\WriteBookmarks\relax
\def\floatpagepagefraction{1}
\def\textpagefraction{.001}
\shorttitle{Soybean yield prediction using transformer-based neural network}
\shortauthors{Li et~al.}
 
\title [mode = title]{SoybeanNet: Transformer-Based Convolutional Neural Network for Soybean Pod Counting from Unmanned Aerial Vehicle (UAV) Images}

\author[1]{Jiajia Li}\ead{lijiajia@msu.edu}
\author[2]{Raju Thada Magar}\ead{thadaraj@msu.edu}
\author[3]{Dong Chen}\ead{chendon9@msu.edu}
\author[2]{Feng Lin}\ead{fenglin@msu.edu}
\author[2]{Dechun Wang}\ead{wangdech@msu.edu}
\author[4]{Xiang Yin}\ead{yinxiang@sjtu.edu.cn}
\author[5]{Weichao Zhuang}\ead{wezhuang@seu.edu.cn}
\author[6]{Zhaojian Li*}\ead{lizhaoj1@egr.msu.edu}

\address[1]{Department of Electrical and Computer Engineering, Michigan State University, East Lansing, MI, USA}
\address[2]{Department of Plant, Soil and Microbial Sciences, Michigan State University, East Lansing, MI, USA}
\address[3]{Environmental Institute \& Link Lab, University of Virginia, Charlottesville, VA, USA}
\address[4]{Department of Automation and Key Laboratory of System Control and Information Processing, Shanghai Jiao Tong University, Shanghai, China}
\address[5]{School of Mechanical Engineering, Southeast University, Nanjing, China}
\address[6]{Department of Mechanical Engineering, Michigan State University, East Lansing, MI 48824, USA}

\address{* Zhaojian Li is the corresponding author}

\begin{abstract}
Soybean is a critical source of food, protein and oil, and thus has received extensive research aimed at enhancing their yield, refining cultivation practices, and advancing soybean breeding techniques. Within this context, soybean pod counting plays an essential role in understanding and optimizing production. Despite recent advancements, the development of a robust pod-counting algorithm capable of performing effectively in real-field conditions remains a significant challenge This paper presents a pioneering work of accurate soybean pod counting utilizing unmanned aerial vehicle (UAV) images captured from actual soybean fields in Michigan, USA. Specifically, this paper presents SoybeanNet, a novel point-based counting network that harnesses powerful transformer backbones for simultaneous soybean pod counting and localization with high accuracy. In addition, a new dataset of UAV-acquired images for soybean pod counting was created and open-sourced, consisting of 113 drone images with more than 260k manually annotated soybean pods captured under natural lighting conditions. Through comprehensive evaluations, SoybeanNet demonstrates superior performance over five state-of-the-art approaches when tested on the collected images. Remarkably, SoybeanNet achieves a counting accuracy of $84.51\%$ when tested on the testing dataset, 
attesting to its efficacy in real-world scenarios. The publication also provides both the source code (\url{https://github.com/JiajiaLi04/Soybean-Pod-Counting-from-UAV-Images}) and the labeled soybean dataset (\url{https://www.kaggle.com/datasets/jiajiali/uav-based-soybean-pod-images}), offering a valuable resource for future research endeavors in soybean pod counting and related fields.

\end{abstract}

\maketitle
\section{Introduction}
\label{sec:intro}
Soybean has been one of the most economically significant leguminous seed crops worldwide. Serving as a primary oilseed crop, they contribute to over twenty-five percent of the global protein supply for both human consumption and animal feed \citep{graham2003legumes}.
As society continues to develop and the global population grows, there is a gradual rise in the demand for soybeans \citep{ray2013yield}. Meeting this growing demand for soybeans is a significant challenge, especially when considering the limitations of available land, agronomic inputs, biotic and abiotic stresses. One of the most efficient strategies to boost soybean production is by enhancing genetic yield gains through plant breeding techniques. This involves the improvement of the average genetic or phenotypic value of a population over the cycle of selection \citep{hazel1942efficiency}. 
Accurate and efficient yield estimation holds greater importance in increasing genetic gain \citep{pedersen2004response}. 
It plays a key role in the decision-making for selecting and developing high-yielding genotypes.
In the research field, soybean yield is determined by the manual weight measurement taken after harvesting. However, this method demands a significant investment of labor, time, and cost, particularly during the early stages of breeding programs when plant breeders need to manage large and diverse populations of soybean plants. Therefore, plant breeders are exploring methods to estimate yield based on highly correlated yield-attributing traits. As such, the number of pods per plant has emerged as a prominent yield-attributing trait in soybeans as genotypes with more pods are likely to produce more seeds and consequently higher yield. 
In this particular context, the development of an efficient method of counting pods and identifying lines with higher pods holds great promise in enabling next-gen soybean breeding with cost-effective and highly accurate yield estimation.




While soybean pod counting is in its early stage of development, plant counting, which bears great similarities,  has gained increased research interest for high-throughput phenotyping by leveraging recent advancements in machine learning and deep learning techniques. These methods have shown promise in various specialty crops, including wheat \citep{xiong2019tasselnetv2, madec2019ear, khaki2022wheatnet}, maize \citep{lu2017tasselnet, liu2020detection},  cotton \citep{xu2018aerial, yadav2023detecting}, among others. Notably, TasselNet \citep{lu2017tasselnet} was proposed for maize tassels counting where a convolutional neural network (CNN) was employed on their collected Maize Tassel Counting (MTC) dataset from four different experimental fields across China between the years 2010 and 2015. These images were collected using a camera mounted on a pole and such data sources offer limited practicality as cameras mounted on poles offer limited throughput, especially when used in expansive agriculture fields. 

To improve data collection efficiency and counting throughput, the use of unmanned aerial vehicles (UAVs) is usually preferred, due to the advantages of a broader field of view, greater flexibility for covering large agricultural areas, and the ability to navigate challenging farm terrains with ease. UAVs also allow continuous monitoring of crop progress, providing insight into genotype and environmental interaction information \cite{tsouros2019review}.
Furthermore, UAVs do not pose risks of damaging plants or causing soil compaction, which can occur when using ground vehicles. 
Over recent years, many researchers have adopted UAVs for plant counting tasks. For example, \cite{liu2020detection} utilized Faster R-CNN \citep{ren2015faster} for the maize tassel detection and counting using UAV images. Similarly, a cotton plant counting framework based on YOLOv3 \citep{redmon2018yolov3} was proposed in \cite{yadav2023detecting} leveraging UAV images. These approaches, akin to some other plant counting works \citep{xu2018aerial, madec2019ear}, are grounded in object detection networks, necessitating the annotation of bounding boxes. However, this bounding box annotation process demands additional labeling time and is often unnecessary for plant counting tasks. Therefore, dot-annotation has become prevalent for plant counting, as exemplified in studies like \cite{lu2017tasselnet, xiong2019tasselnetv2, khaki2022wheatnet}. 


Compared to plant counting, soybean pod counting is generally more challenging due to the small size, high density, and overlapping nature of the pods, especially in the natural fields. Several works have attempted to address the challenges.  For instance, \cite{zhao2023improved} proposed an improved field-based soybean seed counting and localization network by using the dot-annotated labeled data acquired from the ground vehicle based on the P2PNet \citep{song2021rethinking}. In \cite{li2019soybean}, a large-scale soybean seed counting dataset was first constructed using dot annotation, and a two-column CNNs architecture was employed with promising results. However, the soybean seeds were laid out in an ideal setting with a black background, which is very different from the in-field environment. Furthermore, \cite{riera2021deep} developed a DL approach for soybean pod counting to enable genotype seed yield rank prediction using in-field video data collected by a ground robot using a bounding box labeled dataset. Similarly, \cite{lu2022soybean} used a deep neural network combined with a generalized regression neural network (GRNN) to accomplish the soybean pod counting based on the bounding box labeled data on potted soybean plants. Subsequently, \cite{xiang2023yolo} proposed a YOLO X framework \citep{ge2021yolox} to address the dense soybean pod counting based on the bounding-box labeled data in a lab environment. 

Despite the aforementioned progress, several drawbacks persist in these existing soybean pod/seed counting works. Firstly, many practitioners and researchers primarily conduct experiments within controlled artificial environments as in \citep{li2019soybean, xiang2023yolo}. While indoor experiments simplify image processing and advance the understanding of the relationship between soybean pod numbers and breeding, the ultimate goal is to apply these works in real-world field scenarios, which are far more complicated than the artificial environment as a result of complexities stemming from factors like plant density and lighting conditions. Secondly, some of the existing works still rely on bounding-box labeled data for network development \citep{riera2021deep, lu2022soybean, xiang2023yolo}. These bounding boxes are costly and time-consuming to label, and they are unnecessary for the soybean counting task. Additionally, these works expend substantial effort on removing duplicate or split instances detected in congested regions. 
Last but not least, all existing works relied on the ground vehicle or mounted camera for data collection, missing out on the potential benefits of UAV imagery technologies for soybean pod/seed counting.  


To fill the aforementioned gaps, in this paper, we present the first-ever soybean dot-annotated dataset collected using UAVs during the 2022 season in Michigan, USA. We then establish a soybean pod counting framework on this novel dataset. Specifically, inspired by \citep{song2021rethinking, liu2021swin, liang2022end}, a point-based counting network architecture is devised, comprising a transformer-based backbone, a regression branch, and a classification branch along with a Hungarian-based one-to-one matcher. To the best of our knowledge, this represents the pioneering effort in developing soybean pod counting technology utilizing a real-world field dataset acquired via UAVs. This research is anticipated to serve as a valuable reference for future studies aimed at developing machine vision systems for soybean pod counting and related applications. The primary contributions of this paper are summarized as follows.
\begin{itemize}
    \item A novel UAV-acquired image dataset for soybean pod counting was created and open-sourced\footnote{\url{https://www.kaggle.com/datasets/jiajiali/uav-based-soybean-pod-images}}. This dataset comprises 113 drone images with 262,611 manually annotated soybean pods,  all captured under natural lighting conditions in real-field settings.

    \item We present the first work of accurate soybean pod counting utilizing Unmanned Aerial Vehicle (UAV) images, captured from authentic soybean fields in Michigan, USA. Specifically, we develop SoybeanNet, a novel point-based counting framework that harnesses powerful transformer backbones to simultaneously count and pinpoint individual soybean pods.
    
    \item Through comprehensive evaluations, SoybeanNet has demonstrated superior performance over five state-of-the-art approaches when tested on the acquired images. Notably, SoybeanNet achieves an impressive accuracy of $84.51\%$ on 18 test images including 38,444 soybean pods, confirming its efficacy in real-world agricultural conditions. We have open-sourced the code\footnote{\url{https://github.com/JiajiaLi04/Soybean-Pod-Counting-from-UAV-Images}} to facilitate further research and development in this field.
\end{itemize}

The rest of the paper is organized as follows. Section \ref{sec: methods} presents the dataset and related technical details of this study. Experimental results and analysis are presented in Section \ref{sec: experiments} whereas Section \ref{sec: limitation and discussion} presents further discussions and limitations of this work. Finally, concluding remarks and potential future works are provided in Section \ref{sec: sum}. 

\section{Materials and Methods} \label{sec: methods}
In this section, we first introduce our UAV-acquired soybean dataset. We then present technical details on the proposed SoybeanNet, which directly predicts the set of point proposals for soybean pods. Lastly, the evaluation metrics and experimental setups are introduced. 

\subsection{UAV-Acquired Soybean Dataset}
\label{sec: soybean data}
In order to establish a robust deep-learning approach for counting soybean pods, a comprehensive set of Unmanned Aerial Vehicle (UAV) soybean images were collected and meticulously labeled.
The image collection was conducted at the full maturation stage during the 2022 season, which was just before the harvesting when there were no leaves in the plant and pods were visible. 
The RGB images of soybean plants were collected using a portable UAV (i.e., DJI Mavic2) equipped with a high-resolution camera flying at a height of around 13 feet and 53 to 58 degree angle. For the sake of image diversity \citep{lu2020survey}, a total of 67 Advance Yield Trials (AYTs) lines grown from two different soybean fields at the Michigan State University research station in two counties, e.g., MSU extension Lenawee County and Hillsdale County, were used for image acquisition under natural field light conditions. 

\begin{figure*}[!ht]
    \centering
{\includegraphics[width=.8\linewidth]{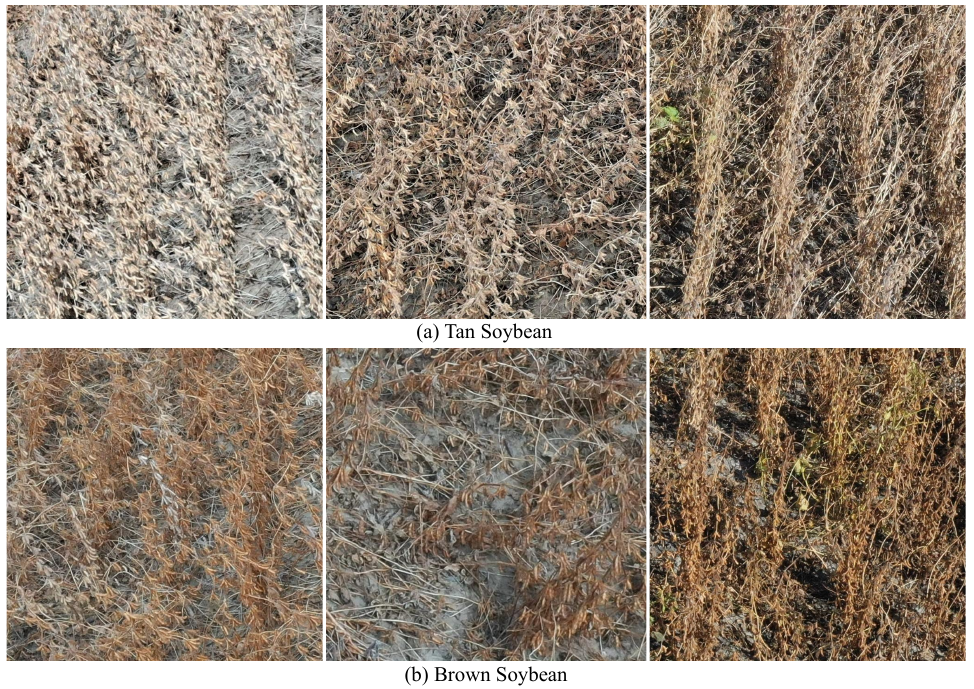} }
    \caption{Illustrations of the test images, comprising 226 images, encompassing both Tan and Brown categories. }
    \label{fig: 2 cls of soybean}
    \vspace{-10pt}
\end{figure*}

In the end, a dataset comprising 113 drone images from the farm of Hillsdale was meticulously annotated, resulting in a total of 262,611 labeled soybean pods. Due to resource limitations for labeling, these images were cropped from the original drone images with a resolution of $5472 \times 3648$, resulting in various resolutions ranging from $800 \times 800$ to $2000 \times 2000$. The images were dot-annotated by labeling the soybean pods as dots at the position of a pod center, using an open-sourced labeling tool named Labelme \citep{Wada_Labelme_Image_Polygonal}. More specifically, the acquired images were first annotated for soybean pods by soybean experts, and the images were then annotated by trained personnel. The final annotations were examined again by the experts to ensure annotation quality. A few unlabeled and labeled sample images from the UAV-acquired soybean dataset are shown in Figure~\ref{fig: 2 cls of soybean} and Figure~\ref{fig: vis} (a) and (c), respectively. It can be seen that these images have large variations in soybean pod color, leaf color, soil background, and field light conditions, which are desirable for establishing DL models that are robust to different soybean classes and natural environments. Furthermore, it is obvious that the soybean pods are generally small, dense, and overlapped, making it particularly challenging for the considered task of soybean pod counting and localization. 
To our knowledge, this marks the inaugural release of an open-sourced UAV-acquired in-field soybean dataset featuring dot annotations. We anticipate that this unique dataset will prove invaluable to the soybean research community, providing a valuable resource to further advance the field of soybean pod counting and related research endeavors.

Furthermore, we set aside 226 test images, comprising 146 images from Hissdale Farm and the remaining 80 images from Lenawee. Based on the soybean pod color the studied genotypes were grouped into two distinct varieties, with 173 images in the Tan category and 53 in the Brown category. Each image is accompanied by their yield values measured in bushels per acre, albeit lacking dot-annotated information, as illustrated in Figure~\ref{fig: 2 cls of soybean}. Moreover, each of these images features approximately four rows of soybean plants, allowing for the calculation of yields in bushels per acre as a viable measurement unit. Even if the yield value is not perfectly accurate, this subset of datasets exhibits greater diversity and complexity compared to the annotated data, which allows us to evaluate the feasibility and generalization of our proposed approach in real-world yield prediction scenarios. 



 

\subsection{SoybeanNet Framework for Soybean Pod Counting}
\label{sec: soybeannet archi}
In this subsection, we introduce SoybeanNet, a novel DL approach for accurate soybean counting. 
Figure~\ref{fig:soybeannet} shows the schematics of SoybeanNet with the tiny backbone version (SoybeanNet-T). The framework consists of three main components: a transformer-based backbone, two branches of regression and classification heads, and a Hungarian algorithm \citep{kuhn1955hungarian} based one-to-one matcher. Given an input image with $N$ soybean pods and the corresponding ground truth labels  $p_i = (x_i, y_i)$, where $i = {1,2,..., N}$ and $(x_i, y_i)$ is the coordinate of the corresponding pod center. The input RGB image $ I \in \mathbb{R}^{H\times W\times 3}$ is first split into non-overlapping patches by a patch-splitting module, where $H$ and $W$ are the height and width of the input image, respectively. 
Then, the split patches are fed into the feature encoding module with a Transformer-based backbone network, outputting two distinct feature maps $\mathcal{F}_2$ and $\mathcal{F}_3$.
Next, the feature map $\mathcal{F}_2$ is added to the upsampled feature map $\mathcal{F}_3$ via the ``neck'' network, utilizing upsampling to enhance the spatial resolution of the feature maps \citep{ronneberger2015u}. 

Subsequently, the resulting feature map $\mathcal{F}_s$ is fed into two parallel branches for point coordinate regression and proposal classification to generate the point proposals $\hat{p}_j$ and the corresponding confidence scores $\hat{c}_j$,  $j = {1,2,..., M}$ with $M$ being the total number of predicted soybean pods. 
The objective is to minimize the distance between the predicted point coordinates $\hat{p}_j$ and their corresponding ground truth coordinates $p_i$, while ensuring a sufficiently high confidence score $\hat{c}_j$.
Note that our framework can simultaneously achieve soybean pod counting and localization. In the following subsections, the major components of SoybeanNet will be explained in more detail.

\subsubsection{SoybeanNet Network Structure}
\label{sec: network}
\begin{figure*}[!ht]
    \centering
{\includegraphics[width=.97\linewidth]{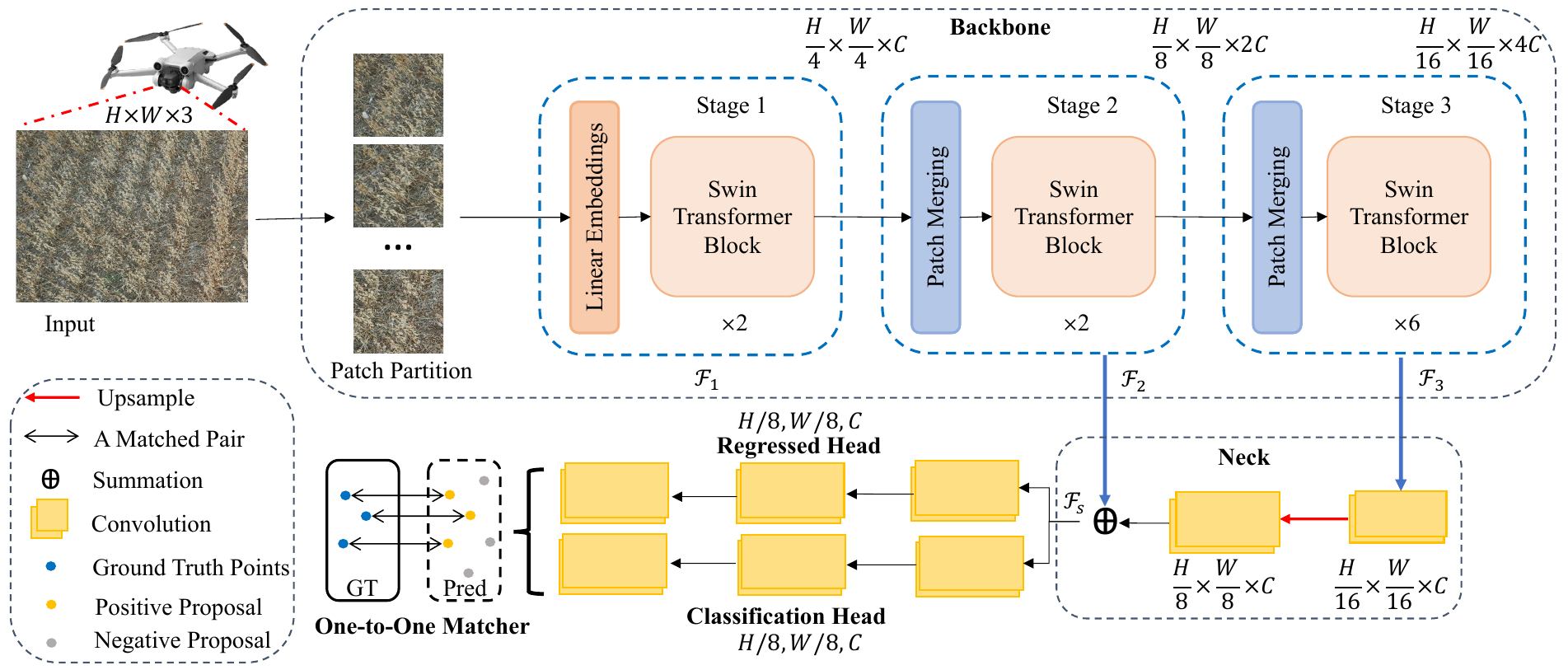} }
    \caption{Architecture of SoybeanNet for soybean pod counting. It is constructed using a transformer-based backbone and gets the deep feature map through the upsampling and skip connection operations. Then a set of predicted point proposals along with the corresponding confidence scores are obtained from two head branches: one for regression and one for classification. The proposed one-to-one matching strategy ensured the alignment between predicted and ground truth points. }
    \label{fig:soybeannet}
    \vspace{-10pt}
\end{figure*}

As depicted in Figure~\ref{fig:soybeannet}, the architectural components of the framework can be categorized into three main modules:  \textbf{backbone}, \textbf{neck}, and \textbf{head}.

\textbf{Backbone}. We employ a transformer-based architecture as our backbone to extract deep features. Transformer models, initially introduced in \cite{vaswani2017attention} and later adapted for image recognition in ViT \citep{dosovitskiy2020image}, have emerged as a groundbreaking neural network paradigm featuring self-attention mechanisms. They have rapidly gained widespread recognition as a versatile backbone, as evidenced by their adoption in various computer vision applications, such as object detection \citep{carion2020end, misra2021end}, image segmentation \citep{strudel2021segmenter}, and crowd counting \citep{sun2021boosting, tian2021cctrans, li2023ccst, gao2022congested, liang2022end}. 
Among these options, Swin Transformers, as described in \citep{liu2021swin}, establish a hierarchical representation by initiating from smaller-sized patches and progressively merging neighboring patches within deeper Transformer layers. This hierarchical feature map structure allows the Swin Transformer model to readily harness advanced techniques like Feature Pyramid Networks (FPN) \citep{lin2017feature} for dense prediction tasks. Notably, it has demonstrated outstanding performance across image classification, object detection, and semantic segmentation tasks, surpassing  ResNet \citep{he2016deep}, ResNeXt \citep{xie2017aggregated}, ViT \citep{dosovitskiy2020image}, and DeiT \citep{touvron2021training} models while maintaining similar computational efficiency on all three tasks. 

Figure~\ref{fig:soybeannet} shows the compact variant (i.e., tiny version) of the transformer-based backbone modified from Swin-Transformer \citep{liu2021swin}. In this architecture, an input RGB UAV image undergoes initial processing through a patch-splitting module, which divides it into non-overlapping patches. Each patch is treated as a ``token'', and its feature representation is constructed by concatenating the raw pixel RGB values. Specifically, a patch size of $4 \times 4$ is utilized, resulting in a feature dimension of $4 \times 4 \times 3 = 48$ for each patch. To further transform these raw-valued features, a linear embedding layer is applied, projecting them into a designated dimension denoted as $C$. The transformer blocks, in conjunction with the linear embedding layer, collectively constitute what is referred to as ``Stage 1'', with a resolution of $\frac{H}{4} \times \frac{W}{4}$. Multiple transformer blocks, featuring modified self-attention computations, are subsequently applied to these patch tokens, thereby constituting ``Stage 2'' and ``Stage 3'', with output resolutions of $\frac{H}{8} \times \frac{W}{8}$ and $\frac{H}{16} \times \frac{W}{16}$, respectively.

We develop a range of Swin Transformer models, each tailored to a specific size and computational complexity requirements. These variants include Swin-Base (Swin-B) as our base backbone model, as well as Swin-Tiny (Swin-T), Swin-Small (Swin-S), and Swin-Large (Swin-L), which represent models of approximately $0.25\times$, $0.5\times$, and $2\times$ of the model size and computational complexity of Swin-B. It is worth noting that the computational complexity of Swin-T and Swin-S is comparable to that of ResNet-50 and ResNet-101 \citep{he2016deep}, respectively. Here are the architectural hyper-parameters for these model variants:
\begin{itemize}
    \item Swin-T: $C = 96$, layer numbers = $\{ 2, 2, 6  \}$;
    \item Swin-S: $C = 96$, layer numbers = $\{ 2, 2, 18 \}$;
    \item Swin-B: $C = 128$, layer numbers = $\{ 2, 2, 18 \}$;
    \item Swin-L: $C = 192$, layer numbers = $\{ 2, 2, 18 \}$.
\end{itemize}

\textbf{Neck}. The architecture's neck component draws inspiration from UNet \citep{ronneberger2015u}. 
We take the feature map $\mathcal{F}_3$ generated by the backbone and increase its spatial resolution by a factor of 2 through nearest-neighbor interpolation \citep{peterson2009k}. Subsequently, this upsampled feature map is merged with another feature map $\mathcal{F}_2$ from a lateral connection through element-wise addition. This lateral connection plays a crucial role in connecting feature maps at different scales and allows the model to capture multi-scale information. Finally, the merged feature map $F_s$ is yielded. 

\textbf{Head}. The head of our framework consists of two branches, both of which are fed with the $F_s$ feature map and generate point localization and confidence scores, respectively. To maintain simplicity, both branches share an identical architecture, comprising three consecutive convolutional layers interspersed with Rectified Linear Unit (ReLU) activations. In the classification branch, it produces confidence scores utilizing Softmax normalization. In the regression branch, it predicts the offsets of the point coordinates, benefiting from the intrinsic translation-invariant property of convolutional layers. This design provides a simple yet effective mechanism for point localization and confidence score estimation within our framework.

\subsubsection{Point Proposal Prediction and One-to-One Matching}
It is noted that multiple proposals may be associated with a single ground truth or vice versa, which can confuse model training and lead to overestimation or underestimation of counts. To address this challenge, we adopt a one-to-one matching strategy in existing works \citep{stewart2016end, carion2020end, song2021rethinking}. This strategy leverages the Hungarian algorithm \citep{kuhn1955hungarian} to establish a bipartite matching between the predicted point proposals and their corresponding ground truth points. Any unmatched proposals are subsequently classified as negative examples (i.e., backgrounds). This approach ensures permutation invariance and guarantees that each target element has a unique corresponding match. 


Let $\Omega( \mathcal{Y}, \hat{\mathcal{Y}}, \mathcal{D})$ be the one-to-one matching strategy, where $\mathcal{Y}$ represents the ground truth point set, $\hat{\mathcal{Y}}$ represents the predicted point proposals, and $\mathcal{D}$ is a pairwise matching cost matrix with dimensions $N \times M$, which quantifies the distance between point pairs, considering both the confidence score and the pixel distance between soybean pods Formally, the cost matrix $\mathcal{D}$ is defined as follows:
\begin{equation} 
\label{equ:D}
\mathcal{D}(\mathcal{Y}, \hat{\mathcal{Y}}) = (\tau \| p_{i} - \hat{p_{j}} \|_{2} - \hat{c}_j)_{i\in N, j\in M},
\end{equation}
where $ \| \cdot \|_{2}$ denotes to the Euclidean distance between the predicted point $\hat{p_{j}}$ and the corresponding ground truth $p_{i}$, and $\hat{c}_j$ is the confidence score of the predicted proposal $\hat{p}_j$. The term $\tau$ acts as a weight factor, balancing the influence of pixel distance and confidence scores. 
This approach guarantees a proficient matching and categorization of point proposals, leading to an improvement in the model's accuracy in counting soybean pods. The identified positive and negative matches serve as the foundation for generating positive proposals and background samples, which are then utilized in the computation of the loss function (as detailed in Section~\ref{sec: loss}).

\subsubsection{Loss Function} \label{sec: loss}
After the one-to-one matching step, the point regression branch and the classification branch in our model are simultaneously optimized. The point localization branch is updated with the regression loss, $\mathcal{L}_{loc}$, defined using the commonly used Euclidean loss:
\begin{equation}
\label{equ:l_loc}
\mathcal{L}_{loc} = \frac{1}{N}\sum_{i=1}^{N} \| p_i - \hat{p}_i \|_{2}^{2},
\end{equation}
where $\left \| \cdot \right \|_{2}$ represents the Euclidean distance. The minimization of this loss term aims at fine-tuning the network's prediction for precise point localization.

The classification branch is updated by minimizing the cross-entropy loss, $\mathcal{L}_{cls}$, defined as:
\begin{equation} 
\label{equ:l_cls}
\mathcal{L}_{cls} = \frac{1}{M}\left \{ \sum_{i}^{N}\log\hat{c}_i + \lambda_{1} \sum_{i = N+1}^{M} \log(1-\hat{c}_i)\right \},
\end{equation}
where $\log\hat{c}_i$ corresponds to the logarithm of the predicted confidence score for positive proposals, while $\log(1-\hat{c}_i)$ represents the logarithm of the complement of the predicted confidence score for negative proposals. The hyperparameter $\lambda_{1}$ is introduced to balance the contributions of the two terms in the classification loss.

The final loss function, denoted as $\mathcal{L}$, is a combination of the regression and classification losses:
\begin{equation}
\label{equ:l}
\mathcal{L} = \mathcal{L}_{loc} + \lambda_{2}\mathcal{L}_{cls}.
\end{equation}
where $\lambda_{2}$ is the weight term for balancing the effect of the classification loss. This composite loss guides the training of our model, ensuring it learns both accurate point localization and effective classification.

\subsection{Evaluation Metrics} \label{sec:eval_metrics}
For evaluating our proposed approach and other state-of-the-art counting methods on our curated dataset, we employ the following evaluation metrics to assess the performance:
\begin{itemize}
    \item Mean Absolute Error (MAE): MAE measures the accuracy of predicted pod counts across the test dataset, indicating how close the predictions are to the ground truth counts.
    \begin{equation} 
    \label{eqn:mae}
    \text{MAE} = \frac{1}{n}\sum_{i=1}^{n} | P_i - G_i |,
    \end{equation}
    where $n$ is the number of test samples,  $P_i$ and $G_i$ represent the estimated number of pods and the actual number of pods in the $i$-th test image, respectively.   
    \item Root Mean Square Error (RMSE): RMSE provides insight into the robustness of the prediction. It measures the square root of the average squared differences between the predicted and actual pod counts as:
    \begin{equation} 
    \label{eqn:rmse}
    \text{RMSE} = \sqrt{\frac{1}{n}\sum_{i=1}^{n}(P_i - G_i)^{2}}.
    \end{equation}
    \item Relative MAE (rMAE): rMAE is often used to normalize MAE with respect to the mean ground truth count. It can be calculated as:
    \begin{equation} 
    \label{eqn:rMAE}
    \text{rMAE} = \frac{1}{n}\sum_{i=1}^{n} \frac{| P_i - G_i |}{G_i}.
    \end{equation}
    \item Accuracy (Acc): accuracy evaluates how well a predictive model aligns with the actual values. 
    \begin{equation} 
    \label{eqn:acc}
    \text{Acc} = 1- \text{rMAE} = 1 - \frac{1}{n}\sum_{i=1}^{n} \frac{| P_i - G_i|}{G_i}.
    \end{equation}
    \item Relative RMSE (rRMSE): rRMSE is often used to normalize RMSE with respect to the mean ground truth count. It can be calculated as:
    \begin{equation} 
    \label{eqn:rRMSE}
    \text{rRMSE} = \sqrt{\frac{1}{n}\sum_{i=1}^{n}\frac{(P_i - G_i)^{2}}{{G_i}^2}}.
    \end{equation}
    \item Coefficient of Determination ($R^2$): $R^2$ measures how well the data fits the trained regression model. It quantifies the proportion of the variance in the predicted counts ($P_i$) that is predictable from the actual counts ($G_i$). A higher $R^2$ value indicates a better fit of the model to the data:
    \begin{equation} 
    \label{eqn:r2}
    \text{$R^2$} = 1 - \frac{\sum_{i=1}^{n}(P_i - G_i)^2}{\sum_{i=1}^{n}(P_i - \bar{G})^2}.
    \end{equation}
    \item Pearson Correlation Coefficient ($r$): $r$ measures the statistical relationship between the predicted number of pods and the actual yields. It quantifies the degree of linear association between the two variables. A positive $r$ value indicates a positive correlation, while a negative value indicates a negative correlation.
    \begin{equation} 
    \label{eqn:R}
    \text{r} = \frac{\sum_{i=1}^{n}(P_i - \bar{P})(G_i - \bar{G})}{\sqrt{\sum_{i=1}^{n}(P_i - \bar{P})^{2}(G_i - \bar{G})^{2}}}.
    \end{equation}
    
\end{itemize}
These evaluation metrics collectively provide a comprehensive assessment of the performance of the counting models, considering accuracy, robustness, and statistical relationships between predicted and actual counts.


\begin{table*}[!h]
\renewcommand{\arraystretch}{1.4}
\centering
\caption{Performance comparisons between SoybeanNet and state-of-the-art baseline methods on the test dataset, where "$\uparrow$" indicates superiority (larger values are better) and "$\downarrow$" indicates preference for lower values.}
\label{tab: benchmarks}
\resizebox{0.8 \textwidth}{!}{%
\begin{tabular}{lccccccc}
\hline
Model        & Venue                                      & MAE ($\downarrow$)   & RMSE $(\downarrow)$   & Acc $(\uparrow)$  & $R^2$  $(\uparrow)$           & rMAE ($\downarrow$)  & rRMSE ($\downarrow$)  \\ \hline  \hline 
Bayesian+        & ICCV'19                                    & 458.71 & 700.33  & 82.21 & 0.7519          & 17.79 & 21.80 \\ 
DM-count         & NeurIPS'20                                    & \underline{393.56} & 546.14  & 82.83 & 0.8204          & 17.17 & 22.65 \\ 
TasselNetV2+     & Front. Plant Sci.'20              & 444.69 & 618.13  & 80.94 & 0.7827          & 19.06 & 22.93 \\ 
M-SFANet & ICPR'21                                    & 678.49 & 1153.00 & 71.55 & 0.4798          & 28.45 & 35.22 \\ 
P2PNet           & ICCV'21                                    & 395.16 & \underline{496.87}  & \underline{82.88} & \underline{0.8615}          & \underline{17.11} & \underline{20.32} \\ \hline
SoybeanNet-T    & \multicolumn{1}{c}{\multirow{4}{*}{Ours}} & 370.71 & 498.58  & 83.90 & 0.8579          & 16.10 & 20.35 \\ 
SoybeanNet-S    & \multicolumn{1}{c}{}                      & \textbf{340.41} & \textbf{449.75}  & 84.50 & \textbf{0.8642} & 16.32 & 20.90 \\  
SoybeanNet-B    & \multicolumn{1}{c}{}                      & 371.75 & 494.51  & 83.53 & 0.8549          & 16.47 & 20.67 \\ 
SoybeanNet-L    & \multicolumn{1}{c}{}                      & 349.93 & 465.71  & \textbf{84.51} & 0.8582          & \textbf{14.89} & \textbf{18.48} \\ \hline
\end{tabular}}
\end{table*}
\subsection{Experimental Setups} \label{sec:eval_setup}
The labeled soybean dataset is randomly partitioned into subsets for training (70\%), validation (15\%), and testing (15\%)  with five different random seeds. Within this partitioning scheme, the training subset serves the purpose of training the soybean-net model, while the validation subset is utilized to select the best-performing model variant. Finally, the testing set remains held out to evaluate the performance of the trained models. The results presented in Table \ref{tab: benchmarks} represent the average performance across these five different random seed partitions on the test dataset.

To enhance the dataset's diversity and robustness, random scaling is applied to the input images, with scaling factors selected from the range [0.7, 1.3]. Importantly, this scaling process ensures that the shorter side of the image remains no smaller than 224 pixels. During the patch partitioning step, the input images are randomly cropped to a fixed size of $224 \times 224$. Additionally, random flipping with a probability of 0.5 is also adopted. These data augmentation techniques contribute to improved model generalization and performance.

We employ a feature map with a stride of $s = 8$ for our predictions.
In the process of matching ground truth and predicted points, we apply a weight term $\tau$ set at $5 \times 10^{-2}$. Within our loss function, $\lambda_{1}$ is given a value of $2 \times 10^{-4}$, and $\lambda_{2}$ is assigned a weight of 0.5. For optimization, we utilize the Adam algorithm \citep{kingma2014adam} with a fixed learning rate of $2 \times 10^{-4}$ and apply a weight decay of $1 \times 10^{-7}$ to regulate the model parameters.

To accelerate the model training process, we employ transfer learning for the backbone architecture. We fine-tune these models using pre-trained weights obtained from the ImageNet-22K dataset \citep{deng2009imagenet}, an extensive dataset consisting of 14.2 million images and 22,000 classes. The training procedure involves using a batch size of 4 over 100 epochs, and we utilize the PyTorch framework (version 1.10.1) \citep{paszke2019pytorch}. Both the training and testing phases of the models take place on a server running Ubuntu 20.04. This server is equipped with two GeForce RTX 2080Ti GPUs, each offering 12GB of GDDR6X memory.

\section{Experiments and Results}
\label{sec: experiments}
In this section, we first introduce the public crowd-counting benchmarks employed in our experiments and proceed to conduct a performance comparison between SoybeanNet and state-of-the-art (SOTA) benchmarks. Subsequently, we illustrate the correlation between the predicted pod count and yield. Finally, we delve into the analysis of the performance across various soybean classes.

\subsection{Comparison with SOTA Benchmarks}
\label{sec: the state-of-the-art}
We conduct an extensive comparison of our method with five state-of-the-art approaches using the soybean dataset. These methods fall into two distinct categories: density map-based methods and localization-based methods.

\begin{figure*}[!h]
    \centering
    \subfigure[Class Tan. Ground Truth Count: 2708 (Left) and Predicted Count: 2532 (Right)]{
    \includegraphics[width=0.99\linewidth]{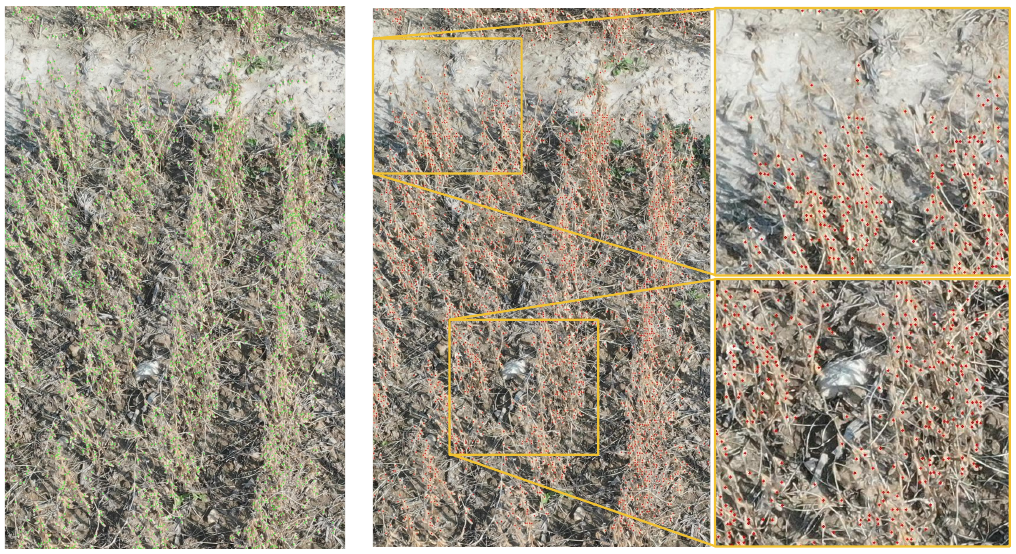}
        \label{fig: sub1}
    }
    \subfigure[Class Brown. Ground Truth Count: 2523 (Left) and Predicted Count: 3505 (Right)]{
    \includegraphics[width=0.99\linewidth]{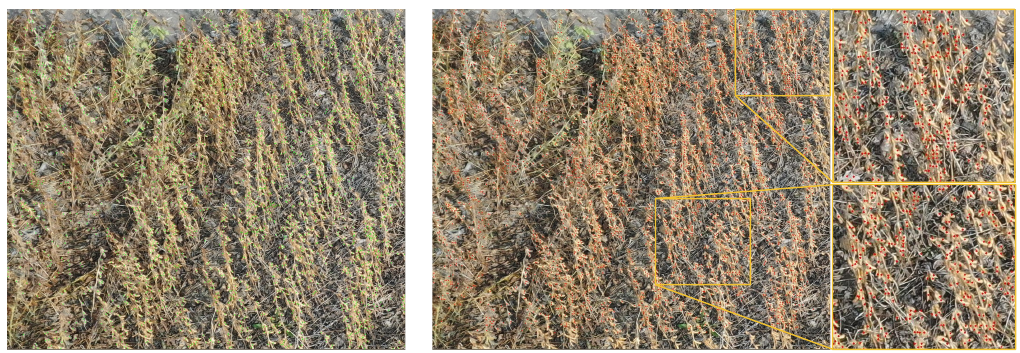}
        \label{fig: sub2}
    }
    \caption{Visualization prediction results for our proposed approach. From left to right, there are images with ground truth points and predicted points, along with the corresponding number of pods. The green points are the manually labeled pods, and the red ones are the predicted points. It can be seen that the proposed method can effectively handle various scenes. Best view via zoom in.}
    \label{fig: vis}
\end{figure*}


\textbf{Density Map based Methods}. Most contemporary methods of crowd-counting adopt the use of density maps, a technique first introduced in \citep{lempitsky2010learning}. These methods estimate the count by summing over the predicted density maps. The following are the notable works in this category:
\begin{itemize}
    \item Bayesian+ \citep{ma2019bayesian}: Bayesian+ introduces a Bayesian loss function to construct a density contribution probability model from point annotations, resulting in significant improvements in crowd-counting tasks.
    \item DM-count \citep{wang2020distribution}: DM-count leverages Distribution Matching to tackle crowd counting, offering an effective approach for this task.
    \item TasselNetV2+ \citep{lu2020tasselnetv2+}: TasselNetV2+ is designed for high-throughput plant counting from high-resolution RGB imagery. It has been evaluated on various plant counting tasks, including wheat ears, maize tassels, and sorghum head counting.
    \item M-SFANet \citep{thanasutives2021encoder}: M-SFANet proposes two encoder-decoder-based architectures that are end-to-end trainable, further advancing the state of the art in crowd counting.
\end{itemize}

\textbf{Localization based Methods}. Localization-based algorithms not only handle counting tasks but also provide precise localization of individual targets within crowds. In this category, the following work represents the SOTA:
\begin{itemize}
    \item P2PNet \citep{song2021rethinking}: P2PNet introduces a purely point-based framework designed for joint counting and individual localization in crowded scenes, offering a comprehensive solution for these tasks.
\end{itemize}

\begin{figure*}[!ht]
    \centering
{\includegraphics[width=.85\linewidth]{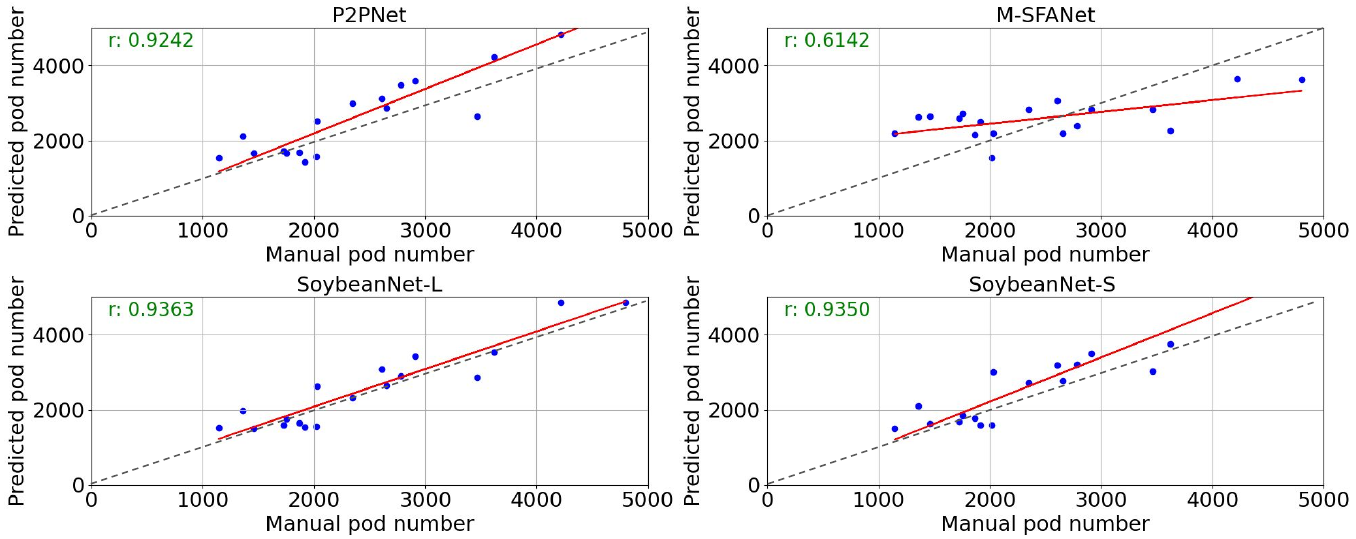} }
    \caption{Comparison between the predicted pod number and manually counted pod number in the test set of soybean data between the best baseline P2PNet, the worst baseline M-SFANet, and the proposed SoybeanNet-L and SoybeanNet-S. The red solid curve is the fitted line and the black dashed curve is the ideal one.}
    \label{fig: r}
    \vspace{-10pt}
\end{figure*}

Table \ref{tab: benchmarks} shows the evaluation results on the SoybeanCount dataset. The top performance is highlighted in bolded numbers and the best benchmark is indicated by underlined numbers. All of the models' backbone is entirely pre-trained using the ImageNet dataset, employing a batch size of 8 for training 100 epochs. Evaluation is conducted on a set of 18 test images with 38,444 soybean pods. The estimated crowd number of our proposed SoybeanNet is obtained by counting the predicted points with confidence scores higher than 0.5. We use MAE and accuracy (i.e., Acc) to evaluate the model accuracy performance and use RMSE, $R^2$, rMAE, and rRMSE to evaluate the model stability. Overall, the proposed SoybeanNet approach outperforms other state-of-the-art counting methods, benefiting from the powerful transformer-based backbone and the effective one-to-one matching algorithm. Specifically, our method outperforms other methods in terms of the accuracy of the model,  achieving a promising MAE of generally less than $370$ and an encouraging Acc of $83.5\%$. In terms of stability, compared with benchmarks, our model is able to achieve an RMSE of $498.58$, an rMAE of $16.47$, an rRMSE of $20.9$, and an $R^2$ of $0.8549$, which demonstrates the effectiveness of our approach on soybean pod counting. In particular, our proposed approach reduces the MAE by $13.5\%$ and MSE by $9.48\%$, as compared to the two best benchmarks, DM-Count \citep{wang2020distribution} and P2PNet \citep{song2021rethinking}, respectively. Remarkably, our methods and P2PNet \citep{song2021rethinking} not only infer the total count number like other benchmarks but also offer the target localization information, which is an advantage for breeding analysis. 

Moreover, it is observed that the performance varies among different backbone choices for our proposed SoybeanNet. 
Specifically, SoybeanNet-B employs the Swin-B backbone, and SoybeanNet-T, SoybeanNet-S, and Soybean-\\Net-L similarly represent the SoybeanNet versions with the Swin-T, Swin-S, and Swin-L backbones, respectively.  Swin-T, Swin-S, and Swin-L correspond to models with approximately $0.25 \times$, $0.5 \times$, and $2 \times$ the model size and computational complexity of the base Swin-B backbone, as elucidated in Section \ref{sec: network}. Overall, the SoybeanNet-S has the best accuracy and robustness. Interestingly, despite SoybeanNet-L's larger size and complexity, it does not achieve the best performance, highlighting that complexity alone does not guarantee superior results. 
%


Figure~\ref{fig: vis} offers a visual representation of our prediction performance on select sample images to illustrate the effectiveness of our approach. The SoybeanNet predicts the center of each soybean pod labeled with a red annotation. The zoom-in areas in both samples include different soybean varieties and different scenes for better viewing. Notably, our SoybeanNet is robust to different soil backgrounds, soybean pod colors, and angles of the soybean plants. Besides, the fitted curve and $r$ between the manual pod number and the predicted count in Figure~\ref{fig: r} demonstrates the effectiveness of our proposed method, representing the highest correlation coefficient $r$ of 0.9353 and 0.9350 achieved by SoybeanNet-L and SoybeanNet-S, respectively. In contrast, M-SFANet yielded the worst performance with a correlation coefficient $r$ of 0.6142.

\subsection{Preliminary Study of the Correlation between Predicted Pod Number and Yield}
The number of pods per plant is a key yield-attributing trait in soybeans and developing cultivars with a higher number of pods per plant is an efficient way to increase genetic yield gains. Several studies have confirmed a strong correlation between the number of pods and yield \citep{lu2022soybean}. In this subsection, we also empirically assess this correlation. Specifically, we employ the Pearson correlation coefficient $r$ to measure the statistical relationship between the predicted number of pods and the actual yields across two distinct fields, Hillsdale County and Lenawee County, as well as with two different soybean categories, Tan and Brown, as illustrated in Figure~\ref{fig: 2 locations, 2 classes}. Our proposed approach achieved a correlation of 0.6618, which showed a strong positive statistical relationship between the predicted pod number and yield. Figure~\ref{fig: 2 locations, 2 classes} (a) illustrates the robustness and generalization of our model. It demonstrates that the performance in Lenawee farm remains consistent with that of Hillsdale trial, despite the fact that our training data exclusively originates from Hillsdale trial.

Regarding the model's performance across different soybean categories, the Tan variety consistently outperforms the Brown variety by $7.21\%$. There are several contributing factors to this discrepancy. Firstly, the issue of unbalanced data plays a significant role; our training dataset predominantly comprises Tan soybeans.
Another factor is the inherent difficulty in recognizing Brown Soybeans. This difficulty stems from their more intricate features and challenges related to overlapping. This complexity is also evident in Figure~\ref{fig: 2 cls of soybean}(b), where certain image regions make it challenging to accurately identify soybean pods. 

It is noteworthy that this is a preliminary experiment. For simplicity, we use the bushels per acre as an approximation of the actual yield for the test images, and all of the test images have approximated four rows of soybean plants. 
Additionally, our labeled data is relatively simple compared to the complexity observed in the set of 236 test images as the labeled data avoids the soybean plants which are 90 degrees from the ground in the view of an image. For instance, our labeled data does not include images resembling those in the third column of Figure~\ref{fig: 2 cls of soybean}. These images exhibit a higher degree of overlapping issues, which we opted to simplify in our annotation process. Despite the aforementioned challenges, the experiments demonstrate the effectiveness of our proposed approach.

\begin{figure*}[!ht]
    \centering
    \subfigure[Two different soybean trials. Hissdale has 146 images and Lenawee has 80 images]{
    \includegraphics[width=0.85\linewidth]{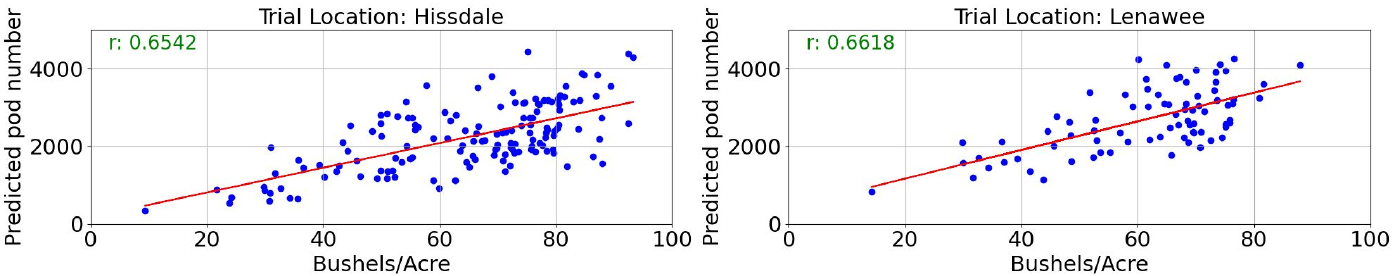}
        \label{fig:subfig1}
    }
    \subfigure[Two classes of soybeans: tan and brown. There are 173 images for the  Tan category and 53 images for the Brown category.]{
    \includegraphics[width=0.85\linewidth]{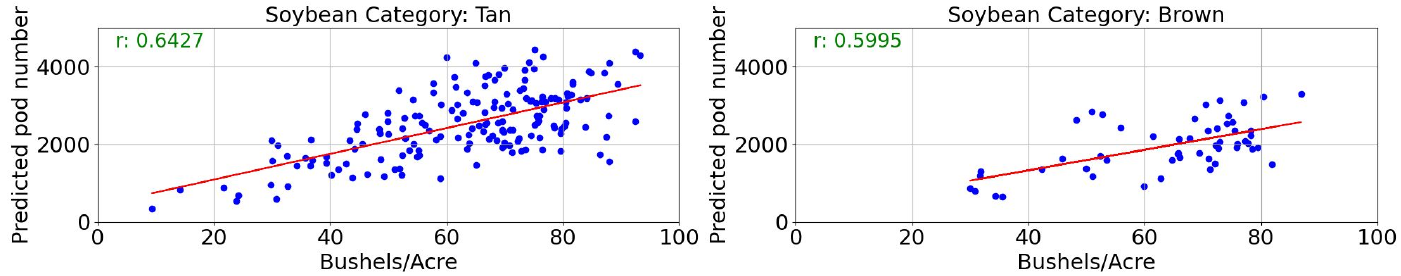}
        \label{fig:subfig2}
    }  
    \caption{The correlation between the predicted pod number and the actual yield on two fields and two classes of soybeans, by using the SoybeanNet-B model.}
    \label{fig: 2 locations, 2 classes}
\end{figure*}

\section{Limitations and Discussions}
\label{sec: limitation and discussion}

While SoybeanNet has demonstrated impressive performance in counting soybean pods within natural environments using UAV-acquired data, there persist several challenges in both the training phase and real-world deployment. These challenges, outlined and discussed below, underscore the importance of comprehending the landscape of opportunities and potential pitfalls in leveraging dense counting for agricultural applications.

\subsection{Resource-Intensive Annotation}
In this paper, we have compiled and made publicly available a soybean dataset comprising 113 drone images. To aid in soybean localization and counting tasks, each soybean pod within the dataset has been meticulously labeled with a dot annotation placed at its center. This comprehensive effort has resulted in the annotation of a total of 262,611 soybean pods, making the dataset a valuable resource for advancing research in soybean analysis and counting. However, it's important to acknowledge the challenges associated with such an extensive manual annotation effort. The manual annotation process incurs substantial costs in terms of time and human resources. Highly skilled annotators are required to maintain consistency and precision in placing dot annotations for each soybean pod. This process, while essential, is labor-intensive and resource-draining, which can limit the scalability of data collection efforts. To overcome these challenges and to enable large-scale data collection, there is a pressing need to explore more efficient annotation methods. 

Recent research on label-efficient learning, as reviewed by our previous work \citep{li2023label}, which harnesses both labeled and unlabeled samples or relies solely on unlabeled samples, holds promise as a solution to mitigate the aforementioned issues. By leveraging the potential of unlabeled data, we can potentially reduce the manual annotation burden while maintaining or even enhancing annotation quality, paving the way for more scalable and cost-effective data collection in agricultural and related domains.

\subsection{Real-Time Deployment}
Throughout this study, our model evaluation was primarily carried out in an offline fashion, involving the pre-collection of images and subsequent testing. This offline evaluation approach has provided valuable insights into the model's performance under controlled conditions, facilitating rigorous assessment and refinement. However, it is worth noting that real-world deployment scenarios often demand real-time or near-real-time inference capabilities, where the model must process data as it arrives. Offline evaluation, while informative, may not capture the challenges and nuances of real-time applications. 

To bridge the gap between offline evaluation and real-time deployment, future research could focus on adapting the model and its infrastructure for real-time use cases. This might involve optimizing the model for inference speed \citep{deng2020model}, exploring edge computing solutions\citep{li2018learning}, and designing efficient data pipelines \citep{um2023fastflow} to feed data to the model in real-time. Additionally, techniques such as model quantization and hardware acceleration \citep{deng2020model} could be explored to enhance inference speed without compromising accuracy.

\section{Conclusion}
\label{sec: sum}
In this paper, we have introduced the pioneering soybean dataset, acquired via UAVs in natural light conditions in Michigan, USA. This dataset comprises 113 drone images, encompassing a remarkable 262,611 annotated soybean pods. To further enhance soybean analysis, we have developed an innovative and efficient soybean pod counting framework named SoybeanNet. This framework leverages powerful transformer-like backbone networks, enabling the concurrent counting and precise localization of individual soybean pods. Our experimental results have demonstrated the superiority of SoybeanNet over state-of-the-art benchmarks. Notably, SoybeanNet-L achieved the highest counting accuracy at an impressive 84.51\% on the testing dataset, while SoybeanNet-S exhibited the best $R^2$ value of 0.8642. Furthermore, a preliminary study shows that our model's predicted pod number has a strong positive statistical relationship with the actual yield. 
To foster research and collaboration, we have made both the source codes for model development and evaluation, as well as the soybean dataset, publicly accessible to the research community. This study not only serves as a foundational resource for informed choices in deep learning models for soybean counting tasks but also holds the potential to benefit research in smart breeding and related fields significantly.

\section*{Authorship Contribution}
\textbf{Jiajia Li}: Conceptualization, Investigation, Software, Writing – Original Draft;
\textbf{Raju Thada Magar}: Conceptualization, Investigation, Software, Writing – Original Draft; 
\textbf{Dong Chen}: Conceptualization, Investigation, Software, Writing – Original Draft; 
\textbf{Feng Lin}: Supervision, Writing - Review \& Editing; 
\textbf{Dechun Wang}: Supervision, Writing - Review \& Editing;  \textbf{Xiang Yin}: Resources, Review \& Editing; 
\textbf{Weichao Zhuang}: Resources, Review \& Editing; 
\textbf{Zhaojian Li}: Supervision, Writing - Review \& Editing.


\typeout{}
\bibliography{ref}
\end{document}